\documentclass[conference]{IEEEtran}
\IEEEoverridecommandlockouts

\usepackage{authblk}
\usepackage{amsmath}
\usepackage{amssymb}
\usepackage{amsfonts}

\usepackage{algorithm}
\usepackage{algorithmic}

\usepackage{booktabs} 
\usepackage[font=small]{caption}
\usepackage{bm}
\usepackage{color}
\usepackage{xcolor}
\usepackage{soul}
\usepackage{arydshln}
\usepackage{inconsolata}
\usepackage{pifont}
\usepackage{graphicx}
\usepackage{colortbl}
\usepackage[comma, sort&compress,square,numbers]{natbib}
\usepackage{hyperref}

\usepackage[table]{xcolor}  
\usepackage{siunitx}  
\definecolor{myred}{RGB}{253,179,179}
\definecolor{myyellow}{RGB}{255,255,185}
\usepackage{tcolorbox}

\usepackage{booktabs}
\usepackage{array}
\usepackage{arydshln}   % load after array/booktabs
\usepackage{adjustbox}
\usepackage{soul}
\usepackage[skip=2pt]{caption}
\usepackage{balance}
\usepackage{subcaption}
\ifCLASSINFOpdf

\else

\fi

\begin{document}

\title{GFix: Perceptually Enhanced Gaussian Splatting Video Compression
\thanks{\noindent \IEEEauthorrefmark{1}Equal contribution. \\ \indent The authors appreciate the funding from the China
Scholarship Council, University of Bristol, Visionular Inc.
and the UKRI MyWorld Strength in Places Programme
(SIPF00006/1).}}

\author{Siyue Teng$^*$\textsuperscript{1}, Ge Gao$^*$\textsuperscript{1}, Duolikun Danier\textsuperscript{2}, Yuxuan Jiang\textsuperscript{1}, Fan Zhang\textsuperscript{1}, Thomas Davis\textsuperscript{3}, Zoe Liu\textsuperscript{3}, David Bull\textsuperscript{1}\\
\textsuperscript{1}\textit{Visual Information Laboratory, University of Bristol, Bristol, BS1 5DD, United Kingdom}\\
\textit{\{siyue.teng, ge1.gao, yuxuan.jiang, fan.zhang, dave.bull\}@bristol.ac.uk}\\
\hspace*{\fill} \textsuperscript{2}\textit{School of Informatics, University of Edinburgh} \hfill \textsuperscript{3}\textit{Visionular Inc., Los Altos, CA 94022 USA} \hspace*{\fill}\\ 
\hspace*{\fill}\textit{duolikun.danier@ed.ac.uk}\hfill \textit {\{thomas, zoeliu\}@visionular.com}\hspace*{\fill}}

\maketitle
\begin{abstract}

3D Gaussian Splatting (3DGS) enhances 3D scene reconstruction through explicit representation and fast rendering, demonstrating potential benefits for various low-level vision tasks, including video compression. However, existing 3DGS-based video codecs generally exhibit more noticeable visual artifacts and relatively low compression ratios. In this paper, we specifically target the perceptual enhancement of 3DGS-based video compression, based on the assumption that artifacts from 3DGS rendering and quantization resemble noisy latents sampled during diffusion training. Building on this premise, we propose a content-adaptive framework, GFix, comprising a streamlined, single-step diffusion model that serves as an off-the-shelf neural enhancer. Moreover, to increase compression efficiency, We propose a modulated LoRA scheme that freezes the low-rank decompositions and modulates the intermediate hidden states, thereby achieving efficient adaptation of the diffusion backbone with highly compressible updates. Experimental results show that GFix delivers strong perceptual quality enhancement, outperforming GSVC with up to 72.1\% BD-rate savings in LPIPS and 21.4\% in FID.

\end{abstract}


\IEEEpeerreviewmaketitle

\section{Introduction}
\label{sec:intro}

Video compression is a key component in  today’s digital communication systems \cite{bull2021intelligent}. While video coding standards such as H.264/AVC \cite{wiegand2003overview} and H.265/HEVC \cite{sullivan2012overview} still dominate the market, research trends have shifted toward learning-based solutions, either by substituting or augmenting specific sub-components with neural networks \cite{laude2016deep, zhang2021video}, or by designing holistic, end-to-end optimizable neural codecs \cite{lu2019dvc,habibian2019video}. Among these end-to-end solutions, scene-agnostic approaches (typically autoencoder-based) \cite{li2021deep, li2024neural} rely on sophisticated architectures to generalize across diverse spatio-temporal content. Despite their impressive compression efficiency \cite{qi2024long}, the associated computational complexity, particularly for decoding, precludes their practical deployment \cite{teng2024benchmarking}. Moreover, these methods primarily target pixel-based distortions, such as MSE, and tend to overlook perceptual aspects that correlate better with human perception.

In contrast, scene-adaptive approaches, such as those based on Implicit Neural Representations (INRs) \cite{sitzmann2020implicit,chen2021nerv,kwan2023hinerv} learn a mapping from coordinates to pixel values, with the quantized network weights serving as the compressed representation. Decoding these scene-specific representations only requires forward passes of compact networks on pixel coordinates, hence resulting in significant complexity reductions. Building on this, recent INR-based approaches \cite{kwan2024nvrc, gao2025givic} have achieved state-of-the-art compression performance whilst also surpassing scene-agnostic counterparts in terms of decoding speed.

More recently, 3D Gaussian Splatting (3DGS) has been widely adopted for neural rendering and reconstruction \cite{kerbl20233d, zhan2025cat}, offering explicit geometric representation with advantages in handling occlusions and complex lighting compared to NeRF methods \cite{mildenhall2021nerf}, despite limitations in transparency and dynamic texture modeling. It has also been extended to 2D image and video compression, delivering promising compression performance \cite{zhang2024gaussianimage,liu2023exploration} and, more importantly, much faster decoding speeds than the INR-based alternatives. However, 3DGS is often more prone to reconstruction artifacts, such as over-smoothing, blob-like structures, and boundary distortions, which become more pronounced under lossy quantization in video compression. These perceptually disruptive artifacts significantly impact the viewing experience, which, coupled with the substantial memory overhead of storing millions of Gaussians, introduces a challenging trade-off between bitrate and reconstruction quality, ultimately placing 3DGS-based codecs at a disadvantage compared to INRs.

Recently, DIFix3D+ \cite{wu2025difix3d+} has demonstrated that images degraded by 3DGS rendering artifacts resemble those at a specific noise level in the forward diffusion process in the latent space (we refer to this as the \textbf{Noise–Artifact Alignment}). This suggests a promising direction for enhancing perceptual quality by directly leveraging pre-trained diffusion models to mitigate 3DGS rendering artifacts. An investigation into whether \ul{the same principle holds for the additional distortions introduced by quantizing and entropy-constraining the Gaussian primitives in video coding} is thus motivated. 

To this end, our paper first \textbf{validates the alignment assumption} in the video compression setting and, building on this insight, proposes a streamlined framework, \textbf{GFix}, which exploits diffusion priors to suppress artifacts with minimal bitrate overhead. Unlike conventional multi-step refinement, our method performs single-step denoising with an \textbf{adaptive, degradation-aware stepsize}, enabling more effective artifact removal whilst maintaining high running efficiency. To further reduce the compression ratio, we introduce \textbf{modulated LoRA (mLoRA)}, a lightweight adaptation technique that reduces the number of trainable parameters and the corresponding bitrate compared to vanilla LoRA \cite{hu2022lora} without quality degradation. Experimental results demonstrate the effectiveness of GFix in enhancing perceptual quality, yielding average BD-rate savings of 72.08\% in terms of LPIPS, relative to the state-of-the-art Gaussian Splatting-based codec, GSVC \cite{liu2023exploration}.

\section{Methodology}
\label{sec:Methodology}

\begin{figure}[!t]
\centering
\small
\begin{minipage}[t]{0.19\textwidth}
  \vspace{0pt}\centering
  \begin{subfigure}[b]{1\linewidth}
      \centerline{\includegraphics[width=3.8cm]{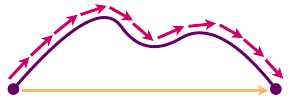}}
      \centerline{Multi/single step diffusion.}
    \end{subfigure}

    \begin{subfigure}[b]{1\linewidth}
      \centerline{\includegraphics[width=3.8cm]{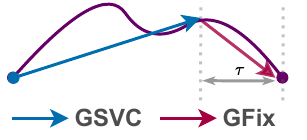}}
      \centerline{GFix with stepsize $\tau$.}
    \end{subfigure}
\end{minipage}%
\hfill
\begin{minipage}[t]{0.29\textwidth}
  \vspace{0pt}\centering
  \includegraphics[width=\linewidth]{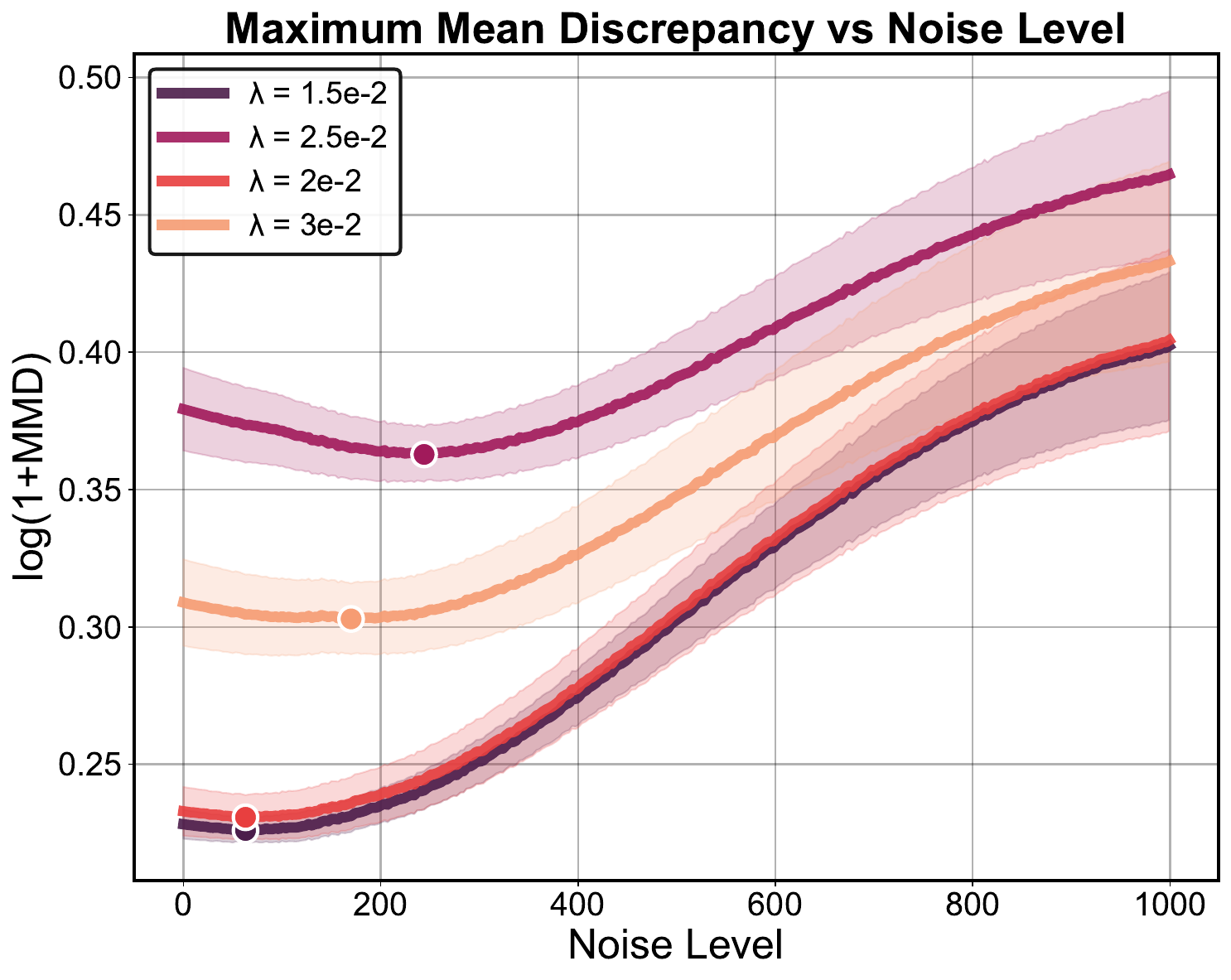}
\end{minipage}

\caption{(\textbf{Left}) Illustration of learnable stepsize. (\textbf{Right}) Average MMD between Gaussian compression artifacts (different compression ratios) and partially noisy images.}
\label{fig:noise_artifact_pair}
\end{figure}

\subsection{Artifact–Noise Alignment in 3DGS Compression} \label{sec:alignment}

Diffusion models \cite{song2021denoising,song2021scorebased} learn data distributions by reversing a gradual noising process, where clean samples are gradually corrupted with Gaussian perturbations following a variance schedule; a neural denoiser (e.g., U-Net) is trained to estimate and remove the noise. DIFix3D+ \cite{wu2025difix3d+} suggests that visual distortions introduced by 3DGS rendering correspond to a particular stage (which we denote as $\tau$) of this forward diffusion trajectory in the latent space. We refer to this effect as \textbf{Noise–Artifact Alignment}. 

\begin{figure}[!t]
  \centering
  \begin{minipage}[t]{\linewidth}
    \vspace{0pt}\centering
    \includegraphics[width=\linewidth]{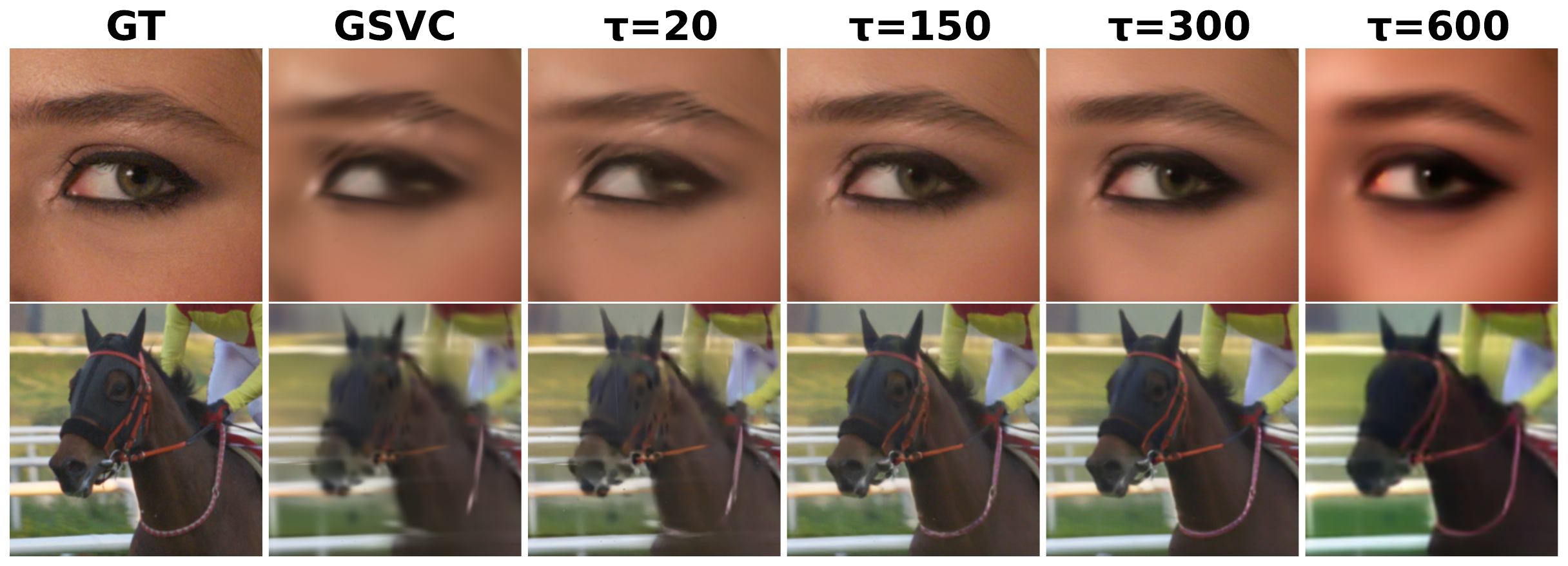}
    \vspace{0.6em}
    {\small
    \resizebox{\linewidth}{!}{%
    \begin{tabular}{rccccccc}
      \toprule
      $\Delta\tau$ & -20 & -15 & -10 & \textbf{Best} & +10 & +20 & +50 \\
      \midrule
      PSNR & 33.21 & \ul{33.54} & 33.08 & \textbf{34.23} & 33.32 & 32.19 & 30.54 \\
      MS-SSIM & 0.975 & 0.974 & \ul{0.968} & \textbf{0.986} & 0.951 & 0.956 & 0.950 \\
      \bottomrule
    \end{tabular}}}
  \end{minipage}
  
  \caption{
    Single-step denoising results at varying noise levels ($\Delta \tau$), 
    with \textbf{visual comparisons} (\textbf{top}) and quantitative metrics (\textbf{bottom}).
  }
  \label{fig:t_compare}
\end{figure}

In the context of video compression, we verify our hypothesis by measuring the Maximum Mean Discrepancy (MMD) between the VAE-encoded latents of GS rendering at different compression ratios and with ground-truth reference latents perturbed with varying noise levels on the UVG dataset. \autoref{fig:noise_artifact_pair} (\textbf{Right}) shows the variation of normalized MMD averaged over the entire dataset; this exhibits a convex profile, with distributional similarity decreasing as the noise level deviates from the optimum. However, no clear correlation between similarity and compression ratio can be observed, likely due to inherent variations in the spatio-temporal complexity of different sequences. The \textit{artifact–noise alignment} is further validated by the improved PSNR and MS-SSIM scores achieved at the MMD-optimal noise level, compared with those obtained at non-optimal levels. This is shown in \autoref{fig:t_compare} (\textbf{bottom}), which reports the average performance at the best noise level for each sequence, along with results for neighboring noise levels (i.e., steps offset by $\Delta \tau$). As illustrated in \autoref{fig:t_compare} (\textbf{top}), non-optimal noise levels either fail to recover fine textures or hallucinate excessive details, leading to noticeable deviations from the original content. These findings substantiate our principle and highlight why diffusion models are particularly effective for mitigating Gaussian Splatting artifacts under lossy compression constraints.

\begin{figure*}[!t]
  \centering
  \includegraphics[width=\linewidth]{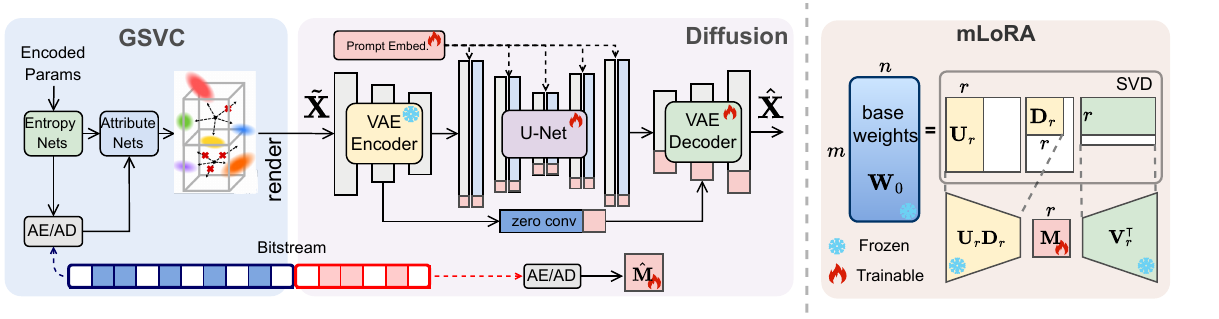}
  \caption{(\textbf{Left}) \textbf{GFix framwork overview}. During decoding, the bitstream is decoded by arithmetic decoding, restoring the reconstructed content of GSVC and the quantized modulation map (based on rounding during inference) $\hat{\mathbf{M}}$. (\textbf{Right}) mLoRA construction.}
  \label{fig: pipeline}
\end{figure*}

\subsection{GFix: The Enhancement Framework}

\autoref{fig: pipeline} illustrates the proposed GFix framework. Given an input video sequence $\mathbf{X} \in \mathbb{R}^{T\times H\times W\times C}$, we first obtain the encoded 3D Gaussian primitives from GSVC \cite{liu2023exploration} (through encoding), a 3DGS-based video codec that represents the video sequence with a set of explicit 3D Gaussian primitives \cite{kerbl20233d}. During decoding, these primitives are first decoded and then projected onto the image plane and rasterized to produce the rendered video frames, $\tilde{\mathbf{X}}$.

Our diffusion model refines $\tilde{\mathbf{X}}$ conditioned on a learnable prompt embedding (with the text prompt initialized as ``\textit{remove degradations}'') to produce the enhanced output $\hat{\mathbf{X}}$.  During encoding, the diffusion model is fine-tuned for only a small number of steps, while the VAE encoder and most decoder layers remain frozen. To enable efficient adaptation, we update only the prompt embedding and a selected subset of U-Net and decoder layers using our proposed mLoRA adapters. These parameters are then entropy coded into a compact form, which is transmitted alongside the original GSVC bitstream.

Based on the discussion in \autoref{sec:alignment}, we further introduce a mechanism to adaptively modulate the denoising strength via a \textbf{learnable stepsize} that adapts to the noise level induced by Gaussian rendering artifacts. As illustrated in \autoref{fig:noise_artifact_pair} (\textbf{Left}), the learnable stepsize enables the diffusion model to automatically align its denoising dynamics with the artifact-induced noise distribution , ensuring optimal restoration after a single denoising step. On the receiver side, these adapter parameters are decoded and used to signal the updates to the diffusion model, which then performs content-adaptive enhancement to $\tilde{\mathbf{X}}$.

\subsection{Modulated LoRA}  

It is noted that fine-tuning the diffusion model in its original, full-rank parameter space, even when limited to just the decoder, is prohibitively expensive. It requires excessive training memory, introduces significant computational overhead and, more critically, produces parameter updates that are extremely costly to compress. Vanilla LoRA \cite{hu2022lora} alleviates this by restricting adaptation to a low-rank subspace, but the trade-off between performance gain (i.e., reduction in distortion) and rank (which correlates to rate) remains unsatisfactory, leaving the low-rank matrices still too large to deliver meaningful improvements in overall compression efficiency.

To address this issue, inspired by recent advancements in the INR literature using modulation for parameter-efficient instance-adaptive overfitting \cite{zhang2024boosting,gao2025pnvc}, we propose a novel \textbf{mLoRA} (\textbf{m}odulated \textbf{LoRA}) technique. For an arbitrary layer with index $i$ to be fine-tuned, we first apply truncated Singular Value Decomposition (SVD) to the base weight $\mathbf{W}^i_0 \in \mathbb{R}^{m \times n}$ (reshaped to 2D) of layer $i$:
\begin{equation}
    \mathbf{W}^i_0 = \mathbf{U}^i \mathbf{D}^i \mathbf{V}^{i \intercal}.
\end{equation} 
The low-rank matrices are initialized as  
$\mathbf{A}^i_r = \mathbf{U}^i_r \mathbf{D}^i_r$ and  
$\mathbf{B}^i_r = \mathbf{V}_r^{i \intercal}$,  
where $r$ denotes the rank, $\mathbf{U}^i_r = \mathbf{U}^i_{[:,\,1:r]} \in \mathbb{R}^{m \times r}$,  
$\mathbf{D}^i_r = \mathbf{D}^i_{[1:r,\,1:r]} \in \mathbb{R}^{r \times r}$, and $\mathbf{V}^i_r = \mathbf{V}^i_{[:,\,1:r]} \in \mathbb{R}^{n \times r}$. Compared to vanilla LoRA \cite{hu2022lora}, instead of overfitting, quantizing, and entropy coding $\mathbf{A}^i_r$ and $\mathbf{B}^i_r$, we keep them frozen and update only a much smaller modulation map $\mathbf{M}^i_r \in \mathbb{R}^{r \times r}$. The updated layer is then parameterized as,
\begin{equation}
\mathbf{W}^{i}_0 \;\rightarrow\; \mathbf{W}^i_0 + \Delta\mathbf{W}^i_0,
\quad \text{where} \quad
\Delta \mathbf{W}^i_0 = \mathbf{A}^i_r \mathbf{M}^i_r \mathbf{B}^i_r.
\end{equation}

\subsection{Quantization and Entropy Modeling}

The mLoRA adapters described above are attached to a total of $N_{\mathrm{FT}}$ layers, including both convolution layers and attention layers. Their modulation maps are aggregated and concatenated across channels.
\begin{equation}
\mathbf{M} = \texttt{concat}\!\left(\mathbf{M}^0_r, \mathbf{M}^1_r, \dots, \mathbf{M}^{N_{\mathrm{FT}}}_r\right) 
\in {\mathbb{R}^{r \times r \times (N_{\mathrm{FT}}+1)}}.
\end{equation}
During training, quantization of $\mathbf{M}$ is simulated with uniform noise $\mathcal{U}(-0.5,0.5)$ along the rate estimation path and with a Straight-Through Estimator (STE) along the distortion path, following prior work \cite{kwan2024nvrc}. For entropy coding, we adopt an empirical non-parametric entropy model \cite{kim2024c3,zhang2024boosting}, which we empirically verified to achieve performance comparable to more sophisticated autoregressive parametric counterparts, likely due to the compactness and high sparsity of the overfitted modulation parameters.

\subsection{Loss function}  
The diffusion backbone is fine-tuned using the following rate-distortion objective:
\begin{equation}
\mathcal{L}_{RD} = R + \lambda D,
\end{equation}
where $R = \mathbb{E}_{p(\tilde{\mathbf{M}})}[-\log_2 q(\tilde{\mathbf{M}})]$ denotes the rate term,  with $p(\tilde{\mathbf{M}})$ representing the true distribution of quantized modulation maps and $q(\tilde{\mathbf{M}})$ as the predicted probability from our entropy model, where $\tilde{\mathbf{M}}$ denotes the noisy approximation of quantized modulation maps during training. The distortion term $D=\lambda_{\text{1}}\mathcal{L}_{\text{LPIPS}}+\lambda_2 \ell_2$ combines perceptual and pixel-wise losses, following DIFix3D+ \cite{wu2025difix3d+}, where $\lambda, \lambda_1, \lambda_2$ are weighting coefficients that balance compression rate and reconstruction fidelity.

\section{Results and Discussion}
\label{sec:experiment}

\subsection{Implementation Details}

Our method builds on SD-Turbo \cite{sauer2024adversarial} as the backbone diffusion model. Input HD frames are divided into non-overlapping $512 \times 512$ patches to mitigate boundary artifacts. The model was trained for 2,000 steps with a batch size of 2, an initial learning rate of 0.05, and a cosine decay schedule. Optimal mLoRA ranks are selected via grid search, yielding 1024 for the prompt embedding, 256 for the U-Net, and 512 for the VAE decoder. For each sequence, five compression levels are obtained using $\lambda \in \{0.03, 0.025, 0.01, 0.005, 0.002\}$.

We conducted our experiments on the widely used UVG dataset \cite{mercat2020uvg}, which contains seven 1080p video sequences. Following \cite{li2023neural}, we used the first 96 frames of each sequence (which corresponds to a total of 672 frames) for evaluation. The sequences were converted from the raw YUV 4:2:0 colorspace to RGB 4:4:4 based on the BT.601 standard.

We compare against one state-of-the-art Gaussian Splatting-based codec, GSVC \cite{liu2023exploration}, one INR-based codec, NeRV \cite{chen2021nerv}, and one conventional codec, H.264 \cite{wiegand2003overview}, under the \textit{medium} preset and \textit{main} profile setting). Results for GSVC and NeRV are reproduced using their original open-source implementations.

Given our focus on perceptual quality \cite{yang2023lossy}, we report VMAF, LPIPS \cite{zhang2018unreasonable} and FID \cite{heusel2017gans}, in addition to the standard PSNR metric. The Bj{\o}ntegaard Delta Rate (BD-rate) \cite{bdrate} of the proposed method against different anchors are adopted to quantify the compression efficiency.

\subsection{Quantitative and Qualitative Analysis}

\begin{figure}[!t]
  \centering
  \includegraphics[width=8.8cm]{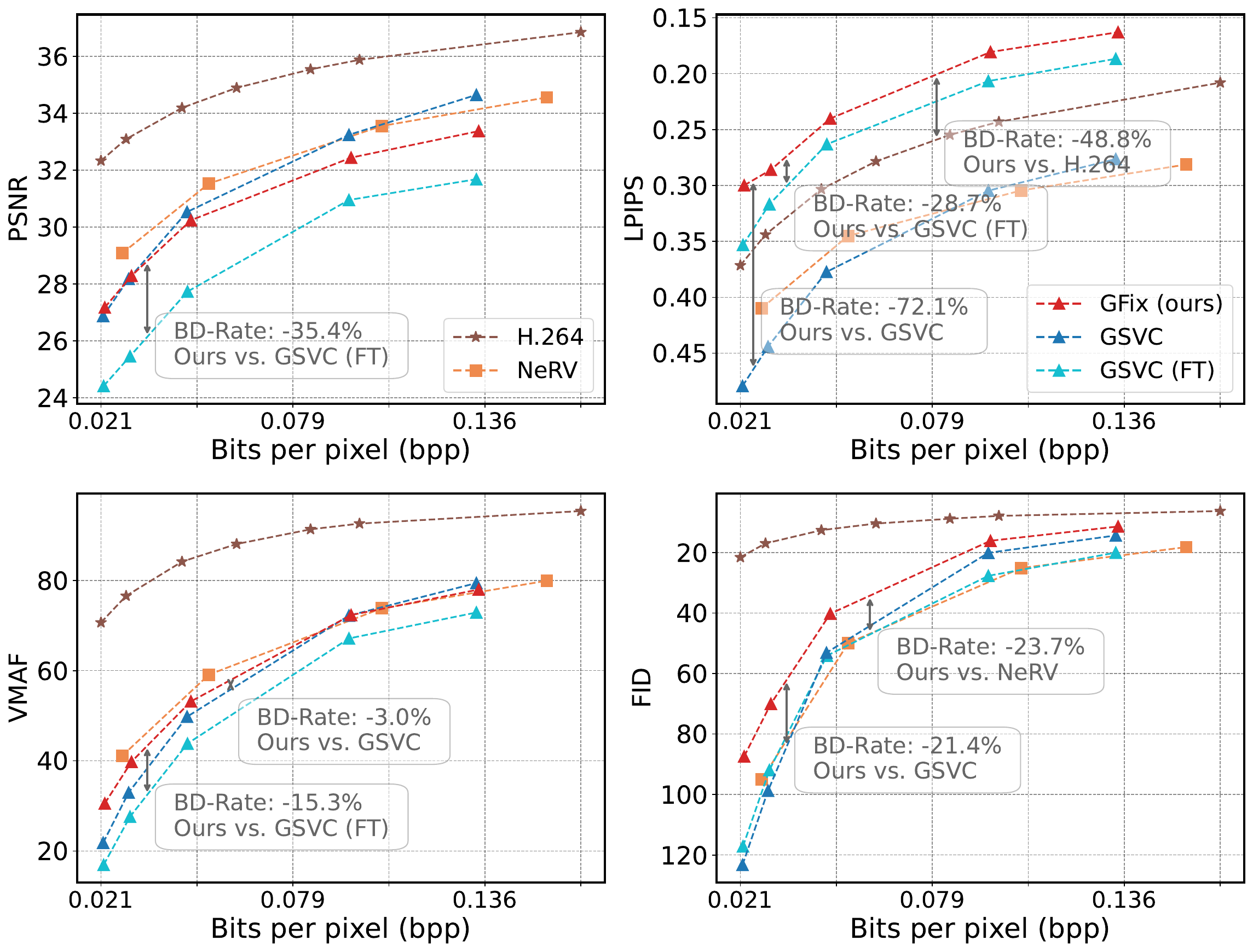}
  \caption{Average rate-quality curves on the UVG dataset. We notice a difference to the reported values of GSVC, which can be attributed to the much shorter sequence length used for evaluation (first 96 frames vs. 600 frames in GSVC).}
  \label{fig:rd-curve}
\end{figure}

\begin{figure}[!t]

\begin{minipage}[b]{1\linewidth}
  \centerline{\includegraphics[width=8.8cm]{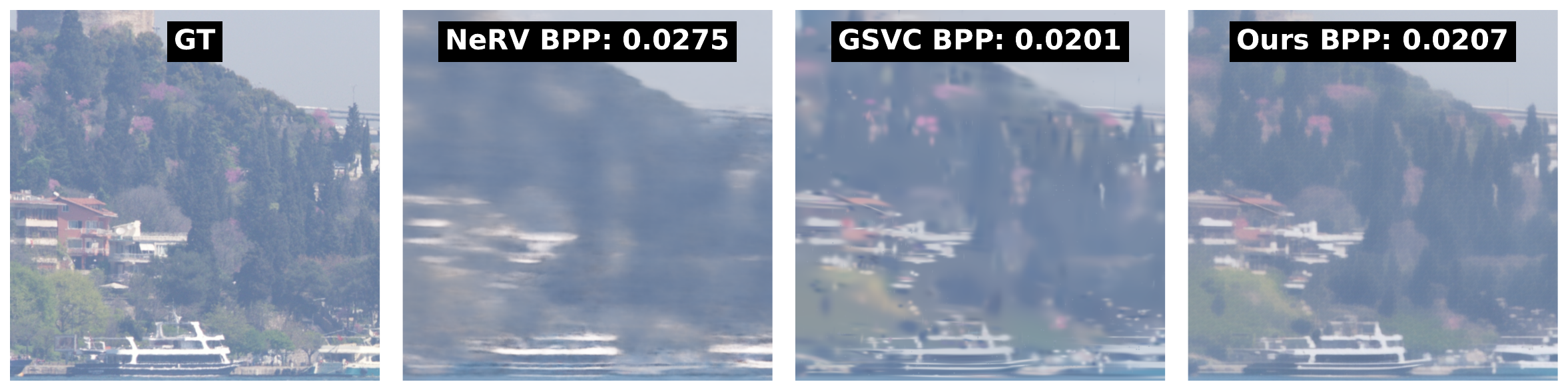}}
%  \vspace{1.5cm}
  % \centerline{(a) ReadySetGo}\medskip
\end{minipage}
% \\
% \begin{minipage}[b]{1\linewidth}
%   \centerline{\includegraphics[width=8.8cm]{img/visual2.pdf}}
% %  \vspace{1.5cm}
%   % \centerline{(b) YachtRide}\medskip
% \end{minipage}
\\
\begin{minipage}[b]{1\linewidth}
  \centerline{\includegraphics[width=8.8cm]{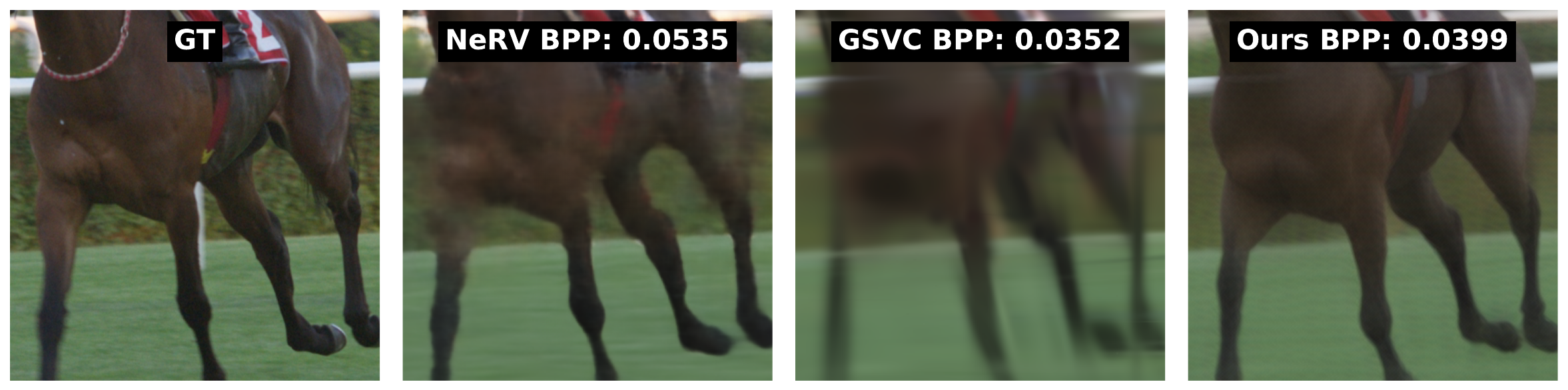}}
%  \vspace{1.5cm}
  % \centerline{(a) ReadySetGo}\medskip
\end{minipage}
\\
\begin{minipage}[b]{1\linewidth}
  \centerline{\includegraphics[width=8.8cm]{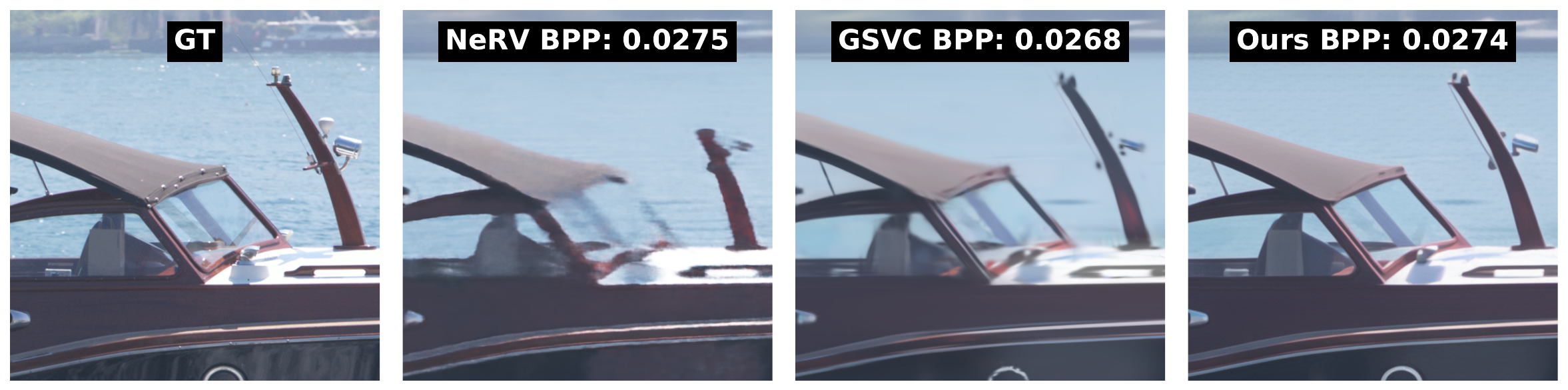}}
%  \vspace{1.5cm}
  % \centerline{(b) YachtRide}\medskip
\end{minipage}
\caption{Visual comparisons of NeRV, GSVC, and the proposed GFix.}
\label{fig:visual-comparison}
\end{figure}

Quantitative results demonstrate the effectiveness of our proposed method across different metrics. As shown in \autoref{fig:rd-curve}, in terms of LPIPS, our approach achieves significant BD-rate savings of 72.1\% and 48.8\% over GSVC and H.264, respectively, and outperforms other codecs across multiple bitrate levels. We also outperform GSVC and NeRV in terms of FID by over 20\%, which further validates the diffusion model's capability in addressing the Gaussian artifacts and improving perceptual quality. With respect to VMAF, our method shows 3.0\% BD-rate improvement over GSVC. Correspondingly, GFix is associated with inferior PSNR performance compared to those distortion-oriented benchmarks, which is aligned with the distortion-perception trade-off theory \cite{blau2018perception}. To ensure a fair comparison, the GSVC baseline (GSVC (FT)) is also fine-tuned with the same number of steps and perceptual loss. The proposed GFix achieves consistent improvements over GSVC (FT) across all evaluation metrics, confirming its effectiveness. The consistent superiority across all perceptual-oriented metrics measured highlight our GFix model's strong perceptual optimization capability. Qualitatively, as shown in \autoref{fig:visual-comparison}, our method achieves significantly better visual quality than GSVC and NeRV, with notably improved detail preservation and a clear reduction in Gaussian and compression artifacts.
\begin{table}[!t]
  \centering
  \caption{Ablation results on the UVG dataset. For v1.x, we only report the reduction in bitstream size (without entropy coding) at a comparable reconstruction quality, as the BD-rate difference would be off scale due to the large size reduction. For v2.x, v2.1 serves as the anchor for BD-rate evaluation.}
  \begin{adjustbox}{width=0.9\linewidth}
  \begin{tabular}{@{}lcc@{}}
    \toprule
    Method & File Size (M) & BD-rate (\%) \\
    \midrule
    (v1.1) LoRA + decoder & 242.04 & -- \\
    (v1.2) mLoRA + decoder & 40.05 & -- \\
    \cdashline{1-3} \addlinespace[0.3ex]
    (v2.1) + entropy model & 0.093 & 0.00\\
    (v2.2)\hspace{0.75em}+ prompt & 0.101 & -6.36 (\textcolor{blue}{6.36$\uparrow$}) \\
    (v2.3)\hspace{1.50em}+ U-Net & 0.171 & -15.61 (\textcolor{blue}{9.25$\uparrow$}) \\
    (v2.4)\hspace{2.25em}+ learnable stepsize & 0.171 & -23.78 (\textcolor{blue}{8.17$\uparrow$}) \\
    \bottomrule
  \end{tabular}
  \end{adjustbox}
  \label{tab:ablation}
\end{table}

\subsection{Ablation Study}
\label{sec:ablation}

To validate the effectiveness of each component, we conducted ablation studies on the first 32 frames of sequence \textit{Beauty, Jockey, ReadySetGo, and YachtRide}. As shown in \autoref{tab:ablation}, our proposed mLoRA achieves a 6-fold reduction in parameter count compared to vanilla LoRA (from 242.04 MB to 40.05 MB) without compression. After incorporating the entropy model to constrain the parameter space during training, the model can be effectively compressed into a compact bitstream of less than 0.01 MB, which only accounts for approximately 7\% of GSVC's original bitstream size. Building upon this entropy-constrained baseline, we observe consistent improvements in BD-rate with each additional component.

\section{Conclusion}
In this paper, we present an entropy-constrained single-step diffusion pipeline with a learnable stepsize for adaptive artifact removal, complemented by a novel modulated LoRA module that improves bitrate compared to the vanilla LoRA, yielding over 6 times the compression ratio while maintaining visual quality. Experimental results on the UVG dataset demonstrate that GFix achieves superior performance against GSVC across all perceptual metrics, with notable improvements over H.264 in LPIPS and NeRV in FID. Future work should explore optimizing GFix for longer sequences through efficient GOP-based coding structures with residual coding and super-resolution \cite{jiang2024mtkd, jiang2025c2d}, which could potentially lead to further improvements in compression efficiency.

\small
\bibliographystyle{IEEEtran}
\bibliography{IEEEexample}

\end{document}